\documentclass[11pt,a4paper]{article}
\usepackage[hyperref]{acl2019}
\usepackage{times}
\usepackage{latexsym}
\usepackage{url}
\usepackage{helvet}  
\usepackage{courier}  
\usepackage{graphicx}  
\usepackage[utf8]{inputenc}
\usepackage{caption}
\usepackage{CJK}
\usepackage{multirow}
\usepackage{amsmath}
\usepackage{amssymb}
\usepackage{color}
\usepackage[title]{appendix}
\usepackage{subcaption}

\aclfinalcopy 
\def\aclpaperid{42} 

\title{Improving Coreference Resolution by Leveraging Entity-Centric Features with Graph Neural Networks and Second-order Inference}

\author{
Lu Liu$^{1}$, Zhenqiao Song$^{1}$, Xiaoqing Zheng$^{1}$ and Jun He$^{2}$  \\
$^1$School of Computer Science, Fudan University, Shanghai, China \\
$^2$Administrative Center of Shanghai R\&D Public Service Platforms \\
{\tt \{luliu19, zqsong17, zhengxq\}@fudan.edu.cn} \\
{\tt jhe@sgst.cn} \\
}

\date{}

\begin{document}
\maketitle
\begin{abstract}
One of the major challenges in coreference resolution is how to make use of entity-level features defined over clusters of mentions rather than mention pairs.
However, coreferent mentions usually spread far apart in an entire text, which makes it extremely difficult to incorporate entity-level features.
We propose a graph neural network-based coreference resolution method that can capture the entity-centric information by encouraging the sharing of features across all mentions that probably refer to the same real-world entity.
Mentions are linked to each other via the edges modeling how likely two linked mentions point to the same entity. 
Modeling by such graphs, the features between mentions can be shared by message passing operations in an entity-centric manner.
A global inference algorithm up to second-order features is also presented to optimally cluster mentions into consistent groups.
Experimental results show our graph neural network-based method combing with the second-order decoding algorithm (named GNNCR) achieved close to state-of-the-art performance on the English CoNLL-2012 Shared Task dataset.
\end{abstract}

\section{Introduction}
\label{sec: intro}
Coreference resolution aims at identifying all the expressions that refer to the same entity in a text. 
It helps to derive the correct interpretation of a text by binding antecedents (or postcedents) with their pronouns together and recognizing the syntactic relationship among them.
The coreference resolution is considered as a critical preprocessing step for various high-level natural language processing (NLP) tasks including document summarization, question answering, and information extraction \cite{chen2016joint,falke2017concept, dhingra2018neural}. 

Existing coreference resolution approaches can be divided into two major categories: mention-pair models \cite{bengtson2008understanding,fernandes2012latent,clark2016deep} and entity-mention models \cite{clark2015entity, wiseman2016learning, end_bert}.
One of the main shortcomings of the mention-pair model is making each coreference decision without entity-level information.
Moreover, the lack of information about the preceding clusters may result in contradictory links.
The entity-mention model tries to make use of the non-local information by encouraging the sharing of features across all mentions that point to the same real-world entity.
However, the coreferent mentions usually spread far apart in a text, which makes it extremely difficult to define effective global features.

\begin{figure}[!t]
    \centering
    \begin{subfigure}{0.15\textwidth}
        \includegraphics[width=\textwidth,height=2.5cm]{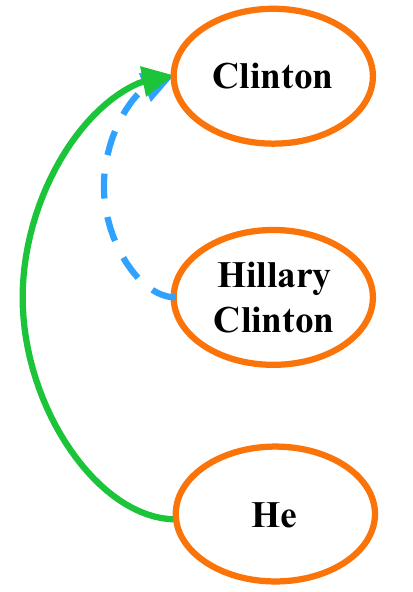}
        \subcaption{}
        \label{1a}
    \end{subfigure}
    \begin{subfigure}{0.15\textwidth}
        \includegraphics[width=\textwidth,height=2.5cm]{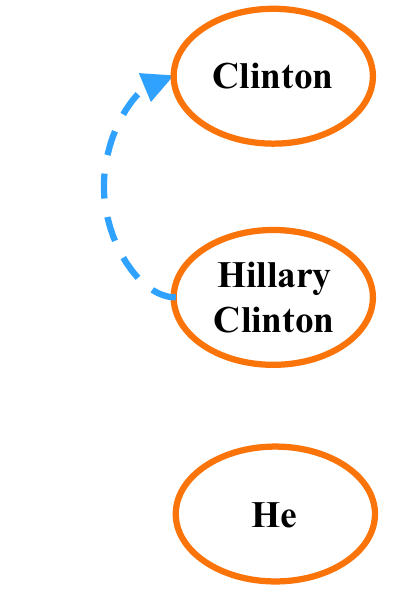}
        \subcaption{}
        \label{1b}
    \end{subfigure}
    \begin{subfigure}{0.15\textwidth}
        \includegraphics[width=\textwidth,height=2.5cm]{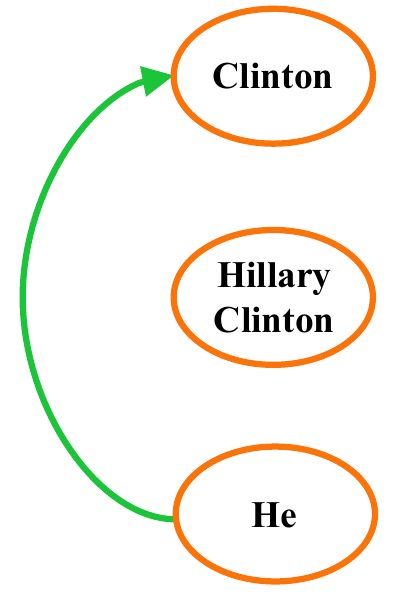}
        \subcaption{}
        \label{1c}
    \end{subfigure}
    \caption{Three different coreference resolution results produce by (a) traditional mention-pair model, (b) entity-centric method but only incorporating the features derived from the preceding clusters, and (c) our graph neural network-based method. They differ in the degree of access to the entity-centric features. The correct arcs are indicated by the green, solid arrows while the incorrect ones by the blue, dashed arrows.}
    \label{figure_intro_1}
\end{figure}

Previous studies either count on the long-term memory (LSTM) or their variants to implicitly capture the global features \cite{end2017, end_biaffine} 
or seek to incorporate the features of the clusters already formed to determine whether a mention is coreferent with a preceding cluster \cite{end_high_order, end_bert}.
The former might miss out some important features for specific pairwise predictions without the help of the explicit entity-level features,
while the latter may suffer from error propagation as false clusters are used to create entity-level features when making future predictions.

Taking the text of ``On November 3, 1992, Clinton was elected the 42nd president of the United States, and the following year Hillary Clinton became the first lady. In 2013, he won the Presidential Medal of Freedom." as an example, we assume that three mentions ``Clinton", ``Hillary Clinton", and ``he" have been well identified.
The traditional mention-pair model is very likely to group these three mentions into a cluster as shown in Figure (\ref{1a}) since ``Clinton" and ``Hillary Clinton" share the same surname, and ``he'' agrees with ``Clinton" both in gender and number.

To make use of information about the clusters already formed, recent studies try to better represent the current mention by incorporating the features derived from the preceding cluster it will most probably join \cite{end_high_order, end_bert}. 
However, those methods only allow such information to be shared in a forward fashion, i.e., from antecedent expressions to postcedent ones, and are prone to reaching the results as shown in Figure (\ref{1a}) and (\ref{1b}). 
The reason is that once ``Hillary Clinton'' is merged with ``Clinton'' to form a cluster, the pronoun ``he'' either joins the formed cluster or begins a new one by itself. 
Even though these errors might be recovered by using a proper decoding algorithm at test time, such as the maximum spanning tree algorithm, similar errors cannot be completely eliminated.

If such information can be shared iteratively in both forward and backward ways, the disagreement in gender between ``Hillary Clinton'' and ``he'' will be detected when the representation of ``Clinton'' is updated by its two possible co-references, which helps to find the correct result as Figure (\ref{1c}). 
Recently, graph neural network (GNN) has gained increasing popularity due to its ability in modeling the dependencies between nodes in a graph \cite{hamaguchi2017knowledge, beck2018graph}. For the coreference resolution, mentions are linked to each other via the edges modeling how likely two linked mentions refer to the same entity. The features between nodes (or mentions) can be shared in each direction with message passing or neighborhood aggregation in an iterative way. We found the entity-centric features can be well captured by GNN, achieving close to state-of-the-art performance.

To avoid contradictory links in mention clustering results, we propose to use a variant of the maximum spanning tree algorithm, second-order decoding algorithm instead of the traditional greedy search algorithm \cite{fernandes2012latent} and the beam search algorithm \cite{bjorkelund2014learning}. 
We factorize the score of a tree into the sum of its arc-pair scores. 
A pair of arcs link three different mentions, and the connected mentions can be viewed as a small cluster. 
Our global inference algorithm up to second-order features helps to define powerful entity-level features between clusters of mentions by aggregating the scores of those small clusters. 

Traditional coreference resolution methods usually include three successive steps: mention detection, candidate pair generation, and mention clustering \cite{haghighi2010coreference, chang2013constrained, clark2016improving}. 
However, recent studies \cite{end2017, end_biaffine, end_high_order} show that joint solutions usually lead to improved performance over pipelined systems by avoiding error propagation.
We follow the line of these research and formulate coreference resolution in a joint manner.

Our contributions are summarized as follows: (1) graph neural networks are introduced to perform coreference resolution, which aims to better leverage the entity-centric information by encouraging the sharing of features across all mentions that refer to the same entity; 
(2) a global inference algorithm up to second-order features is presented to optimally cluster mentions into consistent groups; 
(3) we show our GNN-based method combing with the second-order decoding algorithm achieved close to state-of-the-art performance on the CoNLL-2012 coreference resolution benchmark.

\section{Related Work}
\label{sec: related}
Coreference resolution is a long-standing challenge and one that is essential to accurately interpret a text for the NLP community \cite{ng2010supervised}.
Its approaches can generally be categorized as mention-pair models \cite{bengtson2008understanding, wiseman2015learning} and entity-mention models \cite{clark2015entity, wiseman2016learning}. The former makes each coreference decision independently without taking global information into consideration while the latter addresses the lack of global information by considering whether a mention is coreferent with a cluster.

There are two important design factors for the entity-mention models: how the entity-level features are captured, and how these features can be used properly in mention clustering. Many methods have been proposed to address the first problem. 
\citet{wiseman2016learning} applied RNNs to learn latent, global representations of entity clusters from their mention elements. 
\citet{end_high_order} tried to derive the antecedent distribution from a span-ranking architecture, then iteratively improved the span representations by using the attention mechanism.
\citet{end_bert} proposed to represent an entity approximately by the sum of all possible mentions belonging to the entity set.

As to the second problem, \citet{clark2015entity, clark2016improving} tried to train an incremental coreference system in which each mention starts in its own cluster, and an agent determines whether to merge pairs of clusters or not at each step.
The iterative method that gradually refines the mention representations has achieved decent performance \cite{end_high_order, end_bert}. 
A beam search was also tested to produce the close to optimal coreference result by exploring possible mention clustering states \cite{bjorkelund2014learning}.

Traditional coreference resolution methods usually involve multiple steps, and the errors of the previous step may propagate to any following one. 
To avoid the error propagation, \citet{end2017} firstly designed an end-to-end coreference system that takes every possible span (or a sequence of words) in a document as a candidate mention. Their model is trained to jointly minimize the loss of mention detection and mention clustering. \citet{end_biaffine} improved the system by using a biaffine attention model to estimate the probability of a mention-pair. To reduce the computational cost of the end-to-end system, \citet{end_high_order} introduced a coarse-to-fine approach that incorporates a less accurate but more efficient bilinear factor, which enables more aggressive pruning without hurting accuracy.

It has been well known that the pre-trained language model can bring improvements on multiple NLP tasks including coreference resolution. 
\citet{peters2018deep} tried the word embeddings trained by ELMo for the coreference resolution, and \citet{devlin2018bert} tested the impact of the word embeddings produced by BERT.
More recently, \citet{joshi2019spanbert, joshi2019bert} proposed to replace original LSTM-based encoders with a pre-trained transformer. By designing a pre-training method that can represent and predict spans of text better, their model achieved state-of-the-art results on the CoNLL-2012 Shared Task dataset.

Observing that existing approaches only allow entity-level information to be shared in a forward fashion (i.e., from antecedent expressions to postcedent ones), we introduce the graph neural network to enable such global information between mentions can be shared in both forward and backward ways. Besides, a global decoding algorithm up to second-order features is proposed to optimize the results of mention clustering. 

\section{Methods}
\label{sec: methods}
Based on the joint learning method proposed by \citet{end2017}, we propose our GNNCR which improves their approach in two different aspects: introduce graph neural networks to model the interaction between the mentions by encouraging the feature sharing among them, and design a global inference algorithm up to second-order features for mention clustering.
  
\subsection{Problem Definition}
Following \citet{end2017}, we factorize the problem of coreference resolution into a series of decisions on every possible span for an input document. 
Given a document $D$ of $T$ words, the number of possible text spans is equal to $T(T+1)/2$, and the goal is to find an antecedent $y_{i}$ for each span $i$. 
A set of candidate antecedents for a span $i$ is $\mathcal{Y}_{i}=\left\{ \epsilon, 1, ...,i-1 \right\}$ that includes all the preceding spans and a dummy antecedent denoted as $\epsilon$. A non-dummy antecedent indicates a coreference link between $i$ and $y_{i}$. We use the dummy antecedent in two ways: the span $i$ is not an entity mention, or the span $i$ is an entity mention but is not coreferent with any previous span.

\subsection{Preliminary}
We briefly describe a baseline model \cite{end2017}, denoted as NECR, which does not utilize entity-level features.
Following \citet{end_bert}, the vector representation for each word is composed of a fixed pre-trained word embedding, a feature vector produced by a one-dimensional CNN running over its characters, and a corresponding BERT embedding \cite{devlin2018bert}. 
Taking those word representations as input, a bidirectional LSTM \cite{hochreiter1997long} with attention mechanism \cite{bahdanau2014neural} is first used to represent the boundaries of spans and their head words. 
Then the feature representation $g_{i}$ for each possible span will be generated by concatenating the representations for its boundaries and head word as well as span length.

We perform a coarse-to-fine pruning step before computing the coreference score $s(i,j)$ like \cite{end_high_order}, where $s(i,j)$ denotes a score that reflects how likely the mention span $i$ and $j$ point to the same entity.
To obtain such score, the NECR takes three factors into account: whether span $i$ is a mention, whether span $j$ is a mention, and whether $j$ is an antecedent of $i$ as follows:
\begin{equation} \small
\begin{split}
s(i,j) & = s_{m}(i) + s_{m}(j) + s_{a}(i, j) \\
\label{equ_score}
s_m(i) & = {w}_m^{\top} \cdot \text{FFNN}_{m}({g}_{i}) \\ 
s_a(i,j) & = {w}_a^\top \cdot \text{FFNN}_{a}([{g}_{i}, {g}_{j}, {g}_{i} \circ {g}_{j}, \phi(i, j)])
\end{split}
\end{equation}
where ${w}_m$ and ${w}_a$ are trainable parameters. ``$\cdot$'' denotes dot product, ``$\circ$'' element-wise multiplication, and FFNN a feed-forward neural network. 
The function $\phi(i, j)$ is used to derive the features from the attributes of speaker, genre and distance.

Since its antecedent can not be known in advance for each mention, the objective is to optimize the marginal log-likelihood over all the correct antecedents implied by the gold clustering:
\begin{equation} \small
\begin{split}
L_{base} & = \log \prod_{i=1}^{N} \sum_{\hat{y} \in \mathcal{Y}_{i} \cap \text{GOLD}(i)} P(\hat{y}) \\
P\left( y_{i} \right) & = \frac{e^{s(i,y_{i})}}{\sum_{y^{'} \in \mathcal{Y}_{i} }e^{s(i,y^{'})}}
\end{split}
\end{equation}
where $\text{GOLD}(i)$ is the gold span cluster that the span $i$ belongs to. If span $i$ does not belong to any cluster, we let  $\text{GOLD}(i)=\{ \epsilon \}$.

\subsection{Graph Neural Networks}
The NECR as discussed above can be viewed as a variant of mention-pair model in which the entity-level features can not be well captured, let alone be incorporated. 
\citet{end_bert} tried to refine each mention representation in a cluster by incorporating the features of all preceding mentions in the same cluster. 
Sharing the entity-level information only in a forward fashion can to some extent help to avoid introducing contradictory links, but can not be used to eliminate all the inconsistent links as demonstrated in Figure \ref{figure_intro_1}.
Therefore, we propose to apply graph neural networks (GNNs) to share features among mentions that refer to the same entity in both forward and backward ways. 

\begin{figure}[!t]
  \centering
  \includegraphics[width=7.5cm]{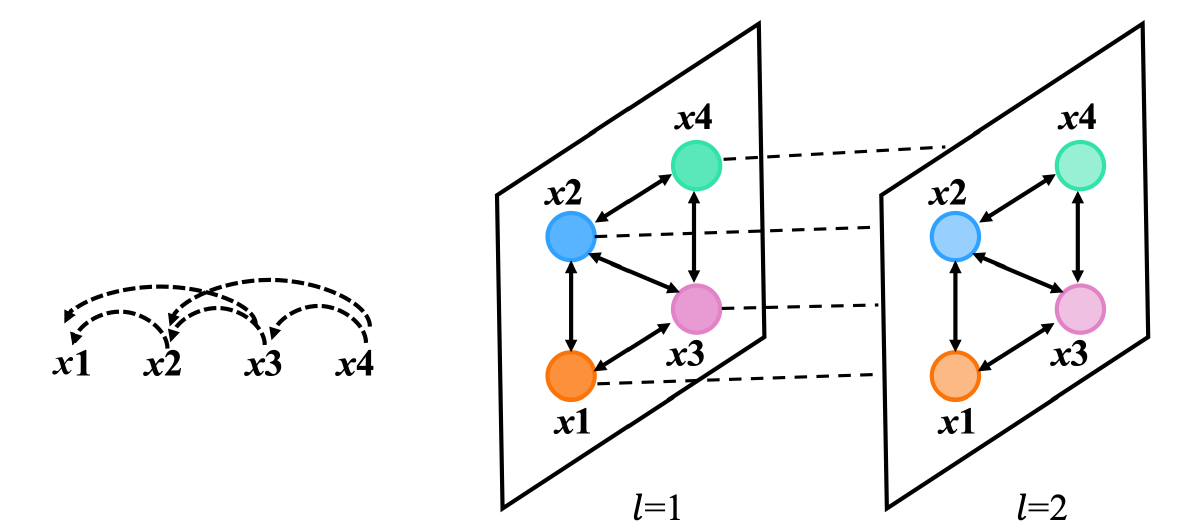}
  \caption{An example of graph neural network.}
  \label{figure_model_1}
\end{figure}
As shown in Figure \ref{figure_model_1}, we represent possible mentions as nodes in a graph, and each mention is connected with its possible antecedents which are chosen through the coarse-to-fine pruning step. A GNN is a multi-layer network, particularly designed to model the interactions among the nodes in a graph. At each layer, the mention representations will be updated by aggregating information from their neighbours according to their similarities. 
Thus, the features across all mentions that refer to the same entity can be shared through multiple-layer updates.

Following \citet{velivckovic2017graph}, the neighbour information is calculated by:
\begin{equation} \small
{a}_{i}^{t} = \sum_{j \in \mathcal{N}(i)} \alpha_{ij}^{t-1} {v}_{j}^{t-1}
\end{equation}
where $\mathcal{N}(i)$ is a set of node $i$'s neighbours, ${v}_{j}^{t-1}$ is the vector representation of node $j$ at the $(t-1)$-th GNN layer (${v}_{j}^{0} = {g}_{j}$), and $\alpha_{ij}^{t-1}$ is the edge weight indicating contribution of node $j$ for building ${a}_{i}^{t}$.

The representation of node $i$ at the $t$-th layer $v_i^t$ is updated through the previous layer vector ${v}_{i}^{t-1}$ and weighted neighbour vector ${a}_{i}^{t}$ by:
\begin{equation} \small
v_i^t = \beta_{i}^{t} \circ a_i^t + (1-\beta_{i}^{t}) \circ v_i^{t-1}
\end{equation}
where $\beta_{i}^{t}$ determines whether to keep the current representation unchanged or to incorporate new information from neighbours. It is updated by:
\begin{equation} \small
\beta_{i}^{t} = \sigma(W_{f}[v_{i}^{t-1}, a_{i}^{t}])
\label{equ_beta}
\end{equation}
where $W_{f}$ is a trainable parameter, and $\sigma$ denotes an activation function.

The edge weights are calculated as follows:
\begin{equation} \small
\alpha_{ij}^{t} = \frac{e^{s({v}_{i}^{t},{v}_{j}^{t})}}{\sum_{y \in \mathcal{Y}_{i} }e^{s({v}_{i}^{t},{v}_{y}^{t})}}
\end{equation}
where $s({v}_{i}^{t},{v}_{y}^{t})$ is the coreference scoring function defined in Equation (\ref{equ_score}). 
At each layer, we use the same scoring function, but feed it with different span representations.
Inspired by \citet{gnn_parsing}, we tried different methods to compute $\alpha_{ij}^{t}$.
One choice is to use a sparse graph that assigns $\left\{ 0, 1\right\}$ values to $\alpha_{ij}^{t}$:
\begin{equation} \small
\alpha_{ij}^{t}=\left\{
\begin{array}{rcl}
1 & & {j=\arg \max_{j^{'}}s({v}_{i}^{t},{v}_{j^{'}}^{t})}\\
0 & & \text{otherwise}\\
\end{array} \right.
\end{equation}
Another alternative is to extend the above method by taking the top-$k$ neighbour nodes into consideration as follows:
\begin{equation} \small
\alpha_{ij}^{t}=\left\{
\begin{array}{rcl}
\frac
{e^{s({v}_{i}^{t},{v}_{j}^{t})}}
{\sum_{y \in \mathcal{N}_{k}^{i}}e^{s({v}_{i}^{t},{v}_{y}^{t})}} & & {j \in \mathcal{N}_{k}^{i}}\\
0 & & \text{otherwise}\\
\end{array} \right.
\end{equation}
where $\mathcal{N}_{k}^{i}$ is a set of nodes with the top-$k$ $s({v}_{i}^{t},{v}_{j}^{t})$ for node $i$. 
In addition, a more trivial approach is also tried, which treats each neighbour equally without using $s({v}_{i}^{t},{v}_{j}^{t})$ by:
\begin{equation} \small
\alpha_{ij}^{t}=\frac{1}{n}, \forall j \in \mathcal{N}(i)
\end{equation}

\subsection{Second-order Decoding Algorithm}
In the NECR, a span $i$ will take span $j$ as its antecedent if the pairwise score $s(i,j)$ is the highest for $i$ comparing to other candidate antecedents. 
Such antecedent prediction made for each possible mention implicitly leads to the clustering results produced by grouping the mentions that are directly or indirectly linked by a series of independent antecedent predictions into a cluster. 
However, it is very likely to cause inconsistent clusters. 
Decoding algorithms, such as the maximum spanning tree, can help to solve this problem.
The main disadvantage of those algorithms is that they can not define the features over any extended scope of the subgraph beyond a single arc.
Therefore, we propose a new decoding algorithm for coreference resolution by introducing a rich feature space.

\begin{figure}[!t]
  \centering
  \includegraphics[width=7.5cm]{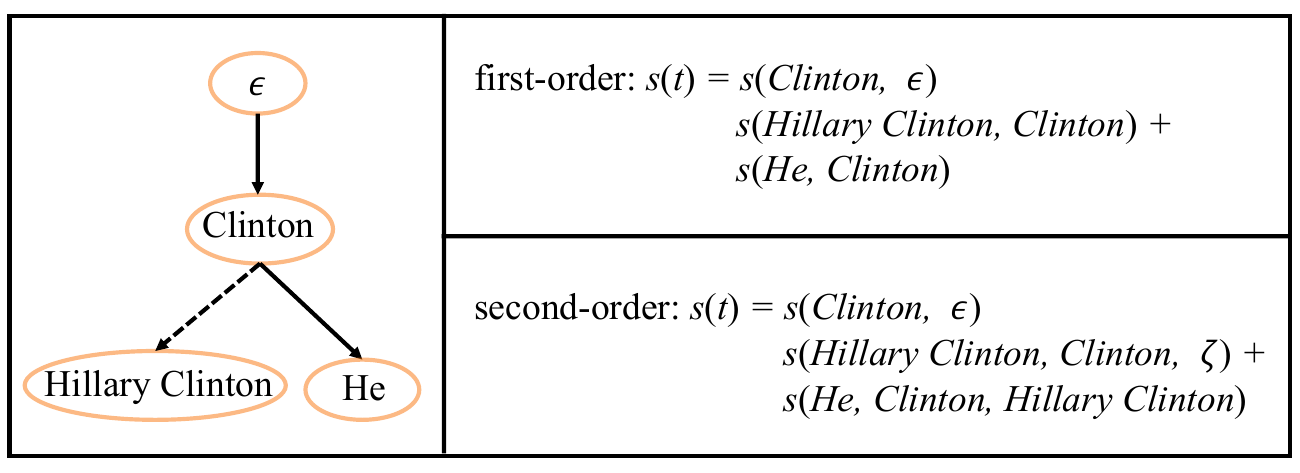}
  \caption{The first-order decoding algorithm cannot use the features over any extended scope of the subgraph beyond a single arc while the second-order one can introduce rich features over the arc-pairs.}
  \label{figure_model_2}
\end{figure}

Unlike the decoding algorithm used by \citet{fernandes2012latent} that views the score of a tree as the summation of independent arc scores (first-order), we factorize the score of a tree into the sum of its arc-pair scores (second-order). 
We do this because simply adding up the score of each single arc without taking the higher-order features into consideration may lead to inferior results.
As shown in Figure \ref{figure_model_2}, a higher score may be given to the left tree by the first-order decoding, which contains a contradictory link. 
However, the second-order decoding will yield a relatively lower score for the arc-pair $s(He, Clinton, Hillary Clinton)$, and the contradictory link will not be created.

The score of an arc-pair $s(i,j,k)$ represents the likelihood that span $j$ is the antecedent of span $i$ and span $k$ is the nearest left sibling of span $i$. If span $i$ is the first child of span $j$, we set $k=\zeta$. When estimating the arc-pair score, we still consider the first-order term to alleviate sparsity problem. Thus, the score can be calculated as follows:
\begin{equation} \small
    s(i,j,k) = \gamma s(i,j) + (1-\gamma) s_{p}(i,j,k)
\end{equation}
where $\gamma$ is a hyperparameter balances the contribution of the first-order score and second-order one. 
The second-order score $s_{p}(i,j,k)$ is calculated by the following formulation:
\begin{equation} \small
s_p(i,j,k) = {w}_p^\top \cdot \text{FFNN}_{p}([{g}_{i}, {g}_{j}, {g}_{k} ])
\end{equation}
where ${w}_p$ is a trainable vector and we set $s_p(i,j,\zeta)$ to $0$.

\begin{table*}[!t]
\begin{center}
\small
\begin{tabular}{l|ccc|ccc|ccc|c}
\hline
\hline
\multirow{2}{*}{Models} & \multicolumn{3}{c|}{MUC} & \multicolumn{3}{c|}{$\rm{B_{3}}$} & \multicolumn{3}{c|}{$\rm{CEAF_{\phi4}}$} &  \\
& Prec. & Rec. & F1 & Prec. & Rec. & F1 & Prec. & Rec. & F1 & Avg. F1 \\
\hline
\cite{wiseman2016learning} & 77.5 & 69.8 & 73.4 & 66.8 & 57.0 & 61.5 & 62.1 & 53.9 & 57.7 & 64.2 \\
\cite{clark2016deep} & 79.2 & 70.4 & 74.6 & 69.9 & 58.0 & 63.4 & 63.5 & 55.5 & 59.2 & 65.7 \\
\cite{clark2016improving} & 79.9 & 69.3 & 74.2 & 71.0 & 56.5 & 63.0 & 63.8 & 54.3 & 58.7 & 65.3 \\
\hline
\cite{end2017} & 78.4 & 73.4 & 75.8 & 68.6 & 61.8 & 65.0 & 62.7 & 59.0 & 60.8 & 67.2 \\
\cite{end_biaffine} & 79.4 & 73.8 & 76.5 & 69.0 & 62.3 & 65.5 & 64.9 & 58.3 & 61.4 & 67.8 \\
\cite{luan2018multi}$^{*}$ & 78.6 & 77.1 & 77.9 & 66.3 & 65.4 & 65.9 & 66.0 & 63.1 & 64.5 & 69.4 \\
\cite{end_high_order}$^{*}$ & 81.4 & 79.5 & 80.4 & 72.2 & 69.5 & 70.8 & 68.2 & 67.1 & 67.6 & 73.0 \\
\cite{end_reinforcement}$^{*}$ & \textbf{85.4} & 77.9 & 81.4 & \textbf{77.9} & 66.4 & 71.7 & 70.6 & 66.3 & 68.4 & 73.8 \\
\cite{end_bert}$^{\dagger}$ & 82.6 & \textbf{84.1} & 83.4 & 73.3 & \textbf{76.1} & 74.7 & 72.4 & 71.1 & 71.8 & 76.6 \\
\cite{joshi2019bert}$^{\S}$ & 84.7 & 82.4 & 83.5 & 76.5 & 74.0 & \textbf{75.3} & \textbf{74.1} & 69.8 & 71.9 & 76.9\\
\hline
NECR $^{\dagger}$ & 82.6 & 83.5 & 83.0 & 73.6 & 75.4 & 74.5 & 71.6 & \textbf{71.6} & 71.6 & 76.4\\
GNNCR $^{\dagger}$ & 84.5 & 83.1 & \textbf{83.8} & 76.2 & 74.1 & 75.1 & 74.0 & 70.5 & \textbf{72.2} & \textbf{77.0}\\
\quad $-$GNN & 84.5 & 82.4 & 83.4 & 76.3 & 73.8  & 75.0 & 73.7 & 70.5 & 72.1 & 76.8\\
\quad $-$Second-order Decoding & 84.4 & 82.8 & 83.6 & 76.2 & 74.0 & 75.1 & 73.7 & 70.6 & 72.1 & 76.9\\
\hline
\hline
\end{tabular}
\end{center}
\caption{Results on the test set of the English CoNLL-2012 shared task. 
The models indicated with $*$ used the word embeddings trained by ELMo \cite{peters2018deep}, 
those indicated with $\dagger$ used BERT \cite{devlin2018bert} embeddings as features,
and the model indicated with $\S$ used BERT as encoder, and fine-tuned parameters using the training data.}
\label{table1}
\end{table*}

The sibling can not be known in advance, thus we maximize the marginal log-likelihood over all correct siblings by:
\begin{equation} \small
\begin{split}
L_{sib} = \log \prod_{i=1}^{N}
\sum_{\hat{j}\in \mathcal{Y}_{i} \cap \text{GOLD}(i),
\hat{k} \in \mathcal{S}_{i,\hat{j}}}
P(\hat{j}, \hat{k}) \\
P\left( \hat{j}, \hat{k} \right) = \frac{e^{s(i,\hat{j}, \hat{k})}}{\sum_{j^{'} \in \mathcal{Y}_{i}, k^{'} \in \mathcal{K}_{i,j^{'}} }e^{s(i,j^{'},k^{'})}}   
\end{split}
\end{equation}
where $\mathcal{K}_{i,j}$ denotes the set of candidate siblings including the dummy sibling $\zeta$ and all spans between span $i$ and $j$. $\mathcal{S}_{i,j}$ denotes a set of gold siblings that are the candidate siblings belonging to the same ground truth cluster as $i$ and $j$.
The overall loss is:
\begin{equation} \small
L = -L_{base}-\lambda L_{sib}
\end{equation}
where $\lambda$ is a hyperparameter governs the relative importance of the first-order term compared with the second-order one. The objective is to find the optimal tree by:
\begin{equation} \small
   \hat{t} = \text{argmax}_{t \in \mathcal{T(A)}} s(t)
\end{equation}
where $\mathcal{T(A)}$ denotes a set of possible trees given the span set $\mathcal{A}$. Following \citet{mcdonald2006discriminative}, we use the ``2-order-non-proj-approx'' algorithm to obtain the optimal tree. The algorithm first applies the second-order Eisner algorithm to get a projective tree, then produces the highest scoring non-projective tree by modifying the projective one.

\section{Experiments}
\label{sec: exp}
\subsection{Implementation Details}
\subsubsection{Dataset and Metrics}
We conducted experiments on the English portion of CONLL-2012 shared task \cite{pradhan2012conll}. 
This corpus contains $2802$ documents for training, $343$ for development, and $348$ for testing. 
Three most popular metrics for coreference resolution were used to evaluate our model: MUC, $\rm{B_3}$ and $\rm{CEAF_{\phi4}}$. 
For each metric, we reported the precision, recall and F1 scores, and took their average F1 score as the final result.

\subsubsection{Hyperparameters}
We used the same hyperparameter settings and optimizer as \cite{end_bert} with the exception that we did not make use of their entity equalization approach to capture entity-level information. In addition, we introduce two new hyperparameters: $\gamma$ and $\lambda$ for applying the second-order decoding algorithm.

Observing that the magnitude of $L_{sib}$ is much larger than $L_{base}$, a relatively small value of $\lambda$ is chosen to balance this difference. Specifically, we tune $\lambda$ in $\{0.01, 0.005, 0.001, 0.0005\}$ according to the average F1 score on the development set, and find that $\lambda = 0.001$ works best.
The hyperparameter $\gamma$ is chosen from $[0, 1]$ with step size $0.1$, where $\gamma=0$ and $\gamma=1$ mean no first-order and second-order terms are employed, respectively. Experimental results indicate that $\gamma=0.8$ performs best among these choices.

\subsection{Empirical Results}
In Table \ref{table1}, we report the results of our method and the models that have achieved a significant improvement on the OntoNotes benchmark over the last three years. 
The first three rows are several representative pipeline models, followed by the recently popular end-to-end ones. 
Our GNNCR and its variants are listed on the bottom part of the table.

Table \ref{table1} shows that our GNNCR significantly outperformed all pipeline models and most end-to-end ones in all cases.
Notably, though the recall score of our GNNCR is a little lower than that of \cite{end_bert}, GNNCR exceeds them on the F$1$ score with a fairly significant margin. It demonstrates that introducing graph neural networks can better leverage entity-centric information than just using the information in a single forward fashion. 
Even if fixed pre-trained BERT features are used, our GNNCR still performs better than \cite{joshi2019bert} which fine-tuned BERT's parameters. It verifies again that the graph structure among entities is helpful to obtain more correct clusters.
\citet{joshi2019spanbert} also propose SpanBERT to better represent and predict spans of text and achieve a new state-of-the-art result (Avg. F1 $79.6$). Although the performance of our GNNCR is a little worse than theirs, it still performs competitively with the fixed pre-trained BERT features.

Ablation tests were also designed to analyse the influence of two components in GNNCR: GNN and the second-order decoding algorithm.
Results show that GNN contributes more to our GNNCR, since it catches entity-level features by aggregating information from all neighbours, while the decoding algorithm mainly catches the nearest sibling features.
Therefore, GNN is capable of capturing richer global information to promote coreference resolution results.
The decoding algorithm is also helpful to improve performance as it forces to eliminate contradictory links with a strong explicit constraint.

As mention detection plays an important role in coreference resolution, we test the performance on this task.
As shown in Table \ref{table2}, our GNNCR achieved the best performance on the F$1$ score.
We also noticed that the precision score of our GNNCR is significantly higher than that of NECR, indicating that introducing entity-level information has the ability to avoid clustering non-referential spans.

\subsection{Model Structure Exploration}
\label{section}
\subsubsection{Graph Neural Networks}
\begin{figure}[!t]
  \centering
  \includegraphics[width=5cm]{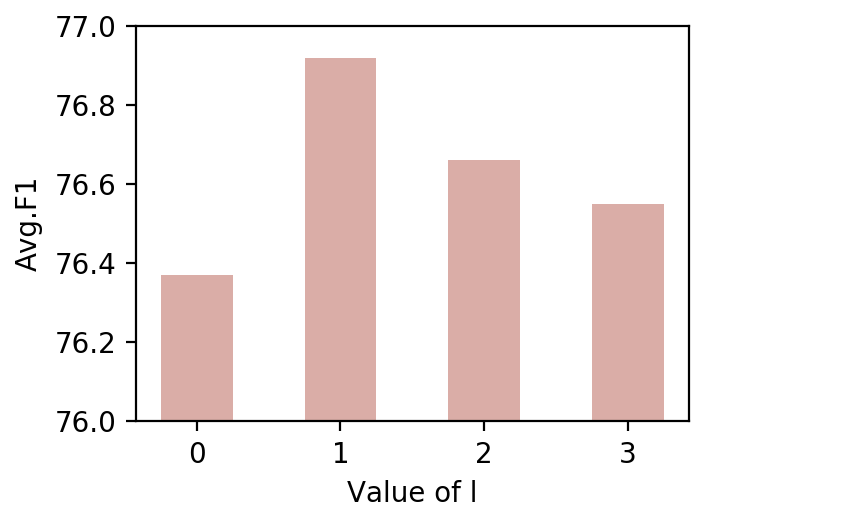}
  \caption{The performance versus the number of layers.}
  \label{layer}
\end{figure}

The influence of different values of layer $l$ is investigated and the results are exhibited in Figure \ref{layer}.
It shows that $1$ layer model significantly outperforms $0$ layer one (NECR) on average F1 score. 
However, with the continuous increase of the layer number, the performance decreased gradually, demonstrating that refining span representations by incorporating features of directly connected nodes is helpful to this task. 
This makes sense because absorbing information from remote nodes may bring noise to mention representations. 
As a result, we set $l=1$ in the following experiments.

\begin{table}[!t]  
\small
\centering
\setlength{\tabcolsep}{4.5mm}{
\begin{tabular}{l|ccc}
\hline
\hline
Model & Prec. & Rec. & F1\\
\hline
\citet{end_high_order} & 86.2 & 83.7 & 84.9 \\
\citet{end_reinforcement} & \textbf{89.6} & 82.2 & 85.7 \\
NECR & 86.9 & \textbf{87.3} & 87.1\\
GNNCR & 88.5 & 86.4 & \textbf{87.4}\\
\hline
\hline
\end{tabular}}
\caption{The mention detection performance reported in precision, recall and F1 score on the test set.}
\label{table2}
\end{table} 

We also test the model performance under different settings of edge weight in GNN.
The results of the unweighted graph (weight all set to 1), hard weight graph (weight renormalized at top-k), and our soft weight graph are shown in Table \ref{table3}.
It can be seen that our soft method surpasses all other competitors.
It is worth noting that using uniform weights would severely hurt the performance, as it does not consider the contributions of different neighbouring nodes.
For hard weight graph-based GNN, the performance is gradually improved with $k$ increasing from $1$ to $3$.

\begin{table}[!htbp]
\small
\centering
\setlength{\tabcolsep}{3mm}{
\begin{tabular}{c|cccc}
\hline
\hline
\multirow{2}{*}{Models} & \multicolumn{4}{c}{Metric F-Scores}\\
& MUC & $\rm{B_{3}}$ & $\rm{CEAF_{\phi4}}$ & Avg.\\
\hline
All = 1 & 83.17 & 74.53 & 71.50 & 76.40\\
Hard-1 & 83.32 & 74.74 & 71.71 & 76.59\\
Hard-2 & 83.40 & 74.70 & 71.80 & 76.63\\
Hard-3 & \textbf{83.68} & 74.87 & 71.88 & 76.81\\
\hline
Soft & 83.58 & \textbf{75.06} & \textbf{72.09} & \textbf{76.91}\\
\hline
\hline
\end{tabular}}
\caption{F1 score on the development set with different settings of edge weights.}
\label{table3}
\end{table}

\begin{table*}[!t] 
\small
\begin{center}
\setlength{\tabcolsep}{3mm}{
\begin{tabular}{l|l}
\hline
\hline
NECR& A: what measures did \textbf{\textcolor[RGB]{225,111,127}{the traffic control department}} eventually take to direct traffic? \\
& B: $\cdots$ \textbf{\textcolor[RGB]{225,111,127}{we}} first set up traffic diversion points for traffic control at the southern ends. \\
& A: $\cdots$ \textbf{\textcolor[RGB]{225,111,127}{the traffic police}} deployed additional manpower on the roads\\
\hline
GNNCR & A: what measures did \textbf{\textcolor[RGB]{225,111,127}{the traffic control department}} eventually take to direct traffic? \\
& B: $\cdots$ \textbf{\textcolor[RGB]{225,111,127}{we}} first set up traffic diversion points for traffic control at the southern ends. \\
& A: $\cdots$ \textbf{\textcolor[RGB]{18,46,186}{the traffic police}} deployed additional manpower on the roads\\
\hline
\hline
NECR & \textbf{\textcolor[RGB]{225,111,127}{Jesus}} knew what \textbf{\textcolor[RGB]{18,46,186}{they}} were thinking. So \textbf{\textcolor[RGB]{225,111,127}{he}} said, ``Why are \textbf{\textcolor[RGB]{18,46,186}{you}} thinking such evil thoughts?" $\cdots$\\
& \textbf{\textcolor[RGB]{225,111,127}{Jesus}} said to 
\textbf{\textcolor[RGB]{18,46,186}{him}}, ``Follow \textbf{\textcolor[RGB]{225,111,127}{me}}." So \textbf{\textcolor[RGB]{18,46,186}{he}} got up and followed \textbf{\textcolor[RGB]{225,111,127}{Jesus}}.\\
\hline
GNNCR & \textbf{\textcolor[RGB]{225,111,127}{Jesus}} knew what \textbf{\textcolor[RGB]{18,46,186}{they}} were thinking. So \textbf{\textcolor[RGB]{225,111,127}{he}} said, ``Why are \textbf{\textcolor[RGB]{18,46,186}{you}} thinking such evil thoughts?" $\cdots$\\
& \textbf{\textcolor[RGB]{225,111,127}{Jesus}} said to 
\textbf{\textcolor[RGB]{55,173,118}{him}}, ``Follow \textbf{\textcolor[RGB]{225,111,127}{me}}." So 
\textbf{\textcolor[RGB]{55,173,118}{he}} got up and followed \textbf{\textcolor[RGB]{225,111,127}{Jesus}}.\\
\hline
\hline
NECR & A: One of the two honorable guests in the studio is Professor \textbf{\textcolor[RGB]{225,111,127}{Zhou Hanhua}} from $\cdots$ \\ 
& A: Next is \textbf{\textcolor[RGB]{18,46,186}{Yang Yang}}, a host of Beijing Traffic Radio Station. $\cdots$ \\
& A: And how \textbf{\textcolor[RGB]{225,111,127}{you}} found out the news on the day of the accident, \textbf{\textcolor[RGB]{225,111,127}{Yang Yang}}?\\
\hline
GNNCR & A: One of the two honorable guests in the studio is Professor \textbf{\textcolor[RGB]{225,111,127}{Zhou Hanhua}} from $\cdots$ \\ 
& A: Next is \textbf{\textcolor[RGB]{18,46,186}{Yang Yang}}, a host of Beijing Traffic Radio Station. $\cdots$ \\
& A: And how \textbf{\textcolor[RGB]{18,46,186}{you}} found out the news on the day of the accident, \textbf{\textcolor[RGB]{18,46,186}{Yang Yang}}?\\
\hline
\hline
\end{tabular}}
\end{center}
\caption{A few example results extracted from the test data. Co-references are highlighted with the same color. We listed three typical examples for which our GNNCR can find correct clusters while the NECR cannot.}
\label{table5}
\end{table*}

\subsubsection{Arc-pair Scoring Function}
We trained a variant of GNNCR which assigned $s(i,j,k)=s(i,j)+s(k,j)+s(i,k)$ for second-order decoding algorithm. 
Results show that the average F1 score of this variant is $1\%$ lower than that of the NECR, indicating that arc-pair information is not a simple linear combination of individual arcs. 
Therefore, it is necessary to learn a new score function $s_p(i,j,k)$ which has the ability to catch sibling relationship.

\subsection{Error Analysis}
In this section, we analyse different types of errors produced by our GNNCR. The analysis tool provided by \citet{kummerfeld2013error} was employed to see which kinds of errors can be relieved with entity-level information.

\citet{kummerfeld2013error} reported the following seven typical error types in coreference resolution: 
(1) Span Error: the detected mention is overlapping with the gold one; 
(2) Missing Entity: an entire entity is missing;
(3) Extra Entity: an entity should be completely removed;
(4) Missing Mention: a mention should be introduced and merged to an entity; 
(5) Extra Mention: a non-referential pronoun is detected; 
(6) Divided Entity: two separated entities should be merged; 
(7) Conflated Entities: the mentions from different clusters are wrongly grouped in the same cluster.

\begin{table}[!htbp] 
\small
\centering
\setlength{\tabcolsep}{4.2mm}{
\begin{tabular}{l|cc}
\hline
\hline
Error &  NECR & GNNCR\\
\hline
Span Error & 275 & \textbf{274} ($-$1)\\
Missing Entity & \textbf{635} & 652 ($+$17)\\
Extra Entity & 464 & \textbf{435} ($-$29)\\
Missing Mention & \textbf{520} & 524 ($+$4)\\
Extra Mention & 592 & \textbf{566} ($-$26)\\
Divided Entity & 1047 & \textbf{1027} ($-$20)\\
Conflated Entities & 974 & \textbf{932} ($-$42)\\
\hline
\hline
\end{tabular}}
\caption{The count of each error type made by GNNCR comparing to the NECR.}
\label{table6}
\end{table} 

The count of each error type produced by GNNCR and NECR is shown in Table \ref{table6}.
It shows that the error count produced by GNNCR are significantly smaller than those by NECR in most types, indicating that leveraging entity-level information is helpful to promote coreference resolution results.
Specifically, compared with NECR, GNNCR achieves lower recall scores (more missing entity and missing mention errors) but relatively higher precision score (less extra entity and extra mention errors). It is because GNNCR is more rigorous when selecting mentions and forming entities. 
Besides, introducing entity-level information does avoid global inconsistency (-42 conflated entities errors).
Overall, the above results demonstrate that our GNNCR has the ability to correctly cluster mentions into consistent groups.

\subsection{Qualitative Analysis}
To gain an insight of how well our GNNCR can integrate the entity-level information, we provide some examples in Table \ref{table5}. 
It can be seen that the clusters generated by our GNNCR can avoid some contradiction.
For example, the NECR assigns ``Professor Zhou Hanhua" as the antecedent of ``you", and ``you" as the antecedent of ``Yang Yang". It seems reasonable when making predictions with local features. 
However, ``Professor Zhou Hanhua" and ``Yang Yang" do not refer to the same entity in a global view. GNNCR can successfully avoid this issue and produce consistent clusters.

\section{Conclusion}
\label{sec: conclusion}
We proposed a coreference resolution system based on graph neural networks and enhanced with the second-order decoding algorithm.
Modeling the mentions and their relationships by the multiple-layer graph neural networks makes it possible to aggregate the features of the mentions pointing to the same entity in an iterative way, while the global inference algorithm up to second-order features helps to produce optimal and consistent clustering results.
Experiments on the English CoNLL-2012 shared task dataset demonstrated that our model achieved close to state-of-the-art performance in the coreference resolution task.

\section*{Acknowledgements}
This work was supported by Shanghai Municipal Science and Technology Project (No. 21511102800).


\bibliography{acl2019}
\bibliographystyle{acl_natbib}
\end{document}